%% file: template.tex
\documentclass{iac}

\begin{document}

\IACconference{76}
\IAClocation{Sydney, Australia}
\IACdates{29 Sep-3 Oct}
\IACyear{2025}
\IACpapernumber{C.1.5.5 (x97477)}
\IACcopyright{2025}{the International Astronautical Federation (IAF)}

\title{Vertical Planetary Landing on Sloped Terrain Using Optical Flow Divergence Estimates}

\IACauthor{Hann Woei Ho}{School of Aerospace Engineering, Universiti Sains Malaysia, Nibong Tebal, Pulau Pinang, Malaysia 14300}{aehannwoei@usm.my}
\IACauthor{Ye Zhou}{School of Aerospace Engineering, Universiti Sains Malaysia, Nibong Tebal, Pulau Pinang, Malaysia 14300}{zhouye@usm.my}

\abstract{Autonomous landing on sloped terrain poses significant challenges for small, lightweight spacecraft, such as rotorcraft and landers. Unlike larger spacecraft, these vehicles have limited processing capability and payload capacity, which makes advanced deep learning methods and heavy sensors impractical. However, flying insects, such as bees, exhibit remarkable landing capabilities despite relying on minimal neural and sensory resources. Their control strategies offer valuable inspiration for spacecraft landing solutions. Flying insects heavily rely on optical flow for navigation and landing. By maintaining a constant flow divergence, a measure of vertical velocity divided by height, they achieve smooth landings where both velocity and height decay exponentially to zero simultaneously. However, adapting this bio-inspired strategy for spacecraft landings on sloped terrain presents two key challenges. First, conventional global estimates of flow divergence average the influence of the sloped surface, which makes them insensitive to terrain inclination. Second, the nonlinearity inherent in flow divergence estimates for landing control can lead to instability when employing conventional linear controllers for spacecraft landing. To address these challenges, this paper proposes a nonlinear control strategy that leverages two distinct local flow divergence estimates to regulate both thrust and attitude during vertical landings. The control law is formulated based on Incremental Nonlinear Dynamic Inversion (INDI) to handle the nonlinear characteristics of flow divergence. The thrust control ensures a smooth vertical descent by keeping a constant average of the local flow divergence estimates, while the attitude control aligns the vehicle with the inclined surface at touchdown by exploiting their difference. To validate the proposed approach, numerical simulations are conducted using a simplified 2D spacecraft model. Multiple landing scenarios are tested with varying surface slopes and flow divergence setpoints to assess the method's robustness and its effect on landing dynamics, respectively. The results demonstrate that by regulating the average flow divergence, the vehicle achieves stable landings, with both velocity and height decreasing exponentially to zero at nearly the same time. In addition, using the difference of the local divergences for attitude control enables effective alignment with the terrain inclination near touchdown, ensuring a safe and controlled landing on inclined surfaces. Overall, this work demonstrates a lightweight and robust landing strategy that improves the feasibility of autonomous planetary missions with resource-constrained vehicles.} 

\maketitle





\section{Introduction}
\label{sec:Intro}
Reliable autonomous flight is essential for extraterrestrial rotorcraft, as the several-minute communication delay between Earth and planetary bodies makes remote piloting infeasible. Therefore, onboard sensing and control systems are critical for enabling safe and effective flight operations. 

NASA's Ingenuity Mars helicopter has demonstrated the feasibility of autonomous flight on Mars, yet it also highlights the challenges of emergency landings on unstructured terrain \cite{nasaIngenuityNews}. Even though some vehicles are equipped with algorithms to select safe landing sites \cite{saldiran2025ensuring, ho2015optical}, unplanned situations may still require landing on non-ideal terrain. In particular, sloped surfaces pose a significant risk: inadequate perception of ground inclination or improper touchdown alignment can lead to instability, vehicle rollover, or rotor damage \cite{yin2025terrain}. These challenges underscore the need for lightweight, robust landing strategies that can ensure controlled touchdown even under resource and sensing constraints.

Flying insects provide an important source of inspiration for lightweight aerial vehicles. Despite their tiny neural and sensory capacity, they perform highly agile flight maneuvers and precise landings in cluttered environments. A key element of their strategy is the use of visual information, particularly optical flow, as the main sensory cue for stabilization and navigation \cite{collett2002memory, srinivasan2000honeybee}. 

Optical flow captures the apparent motion of features in the visual scene, from which insects can extract cues for hovering \cite{ho2024optical}, obstacle avoidance \cite{hu2025seeing, ho2024frontal}, and landing \cite{xie2024bio, ho2024optical, ho2013automatic}. One specific observable, the flow divergence, defined as the ratio of vertical velocity to height \cite{ho2016characterization}, has been shown to play a crucial role in insect landings. By regulating flow divergence to a constant value, insects achieve smooth landings, where both height and velocity decay exponentially to zero at the same time \cite{ho2024optical}. Inspired by these biological strategies, optical flow–based methods have been extensively studied for Micro Aerial Vehicles (MAVs) and offer a lightweight alternative for spacecraft, where payload and computational resources are similarly constrained.

However, exploiting optical flow observables for control introduces significant challenges due to their nonlinear dependence on vehicle states. Classical control strategies, such as PID controllers, have been widely applied to track optical flow for stabilization and landing \cite{ho2013automatic}. However, their linear nature limits stability and performance to specific operating conditions. Gain-scheduled variants attempt to address these limitations \cite{kendoul2014four}, but require careful parameter tuning and accurate models, which are often unavailable in real-world settings. 

Learning-based approaches offer greater performance, yet they introduce heavy computational demands and often depend on pre-training \cite{kooi2021inclined, eschmann2024learning}, which makes them unsuitable for planetary vehicles. Nonlinear control frameworks provide an alternative, with Nonlinear Dynamic Inversion (NDI) establishing a direct input–output relationship \cite{slotine1991applied}. However, conventional NDI relies on precise models, which are rarely available in practice. To overcome this, Incremental Nonlinear Dynamic Inversion (INDI) was developed, leveraging real-time sensor feedback to compensate for model inaccuracies while retaining the efficiency of NDI \cite{sieberling2010robust, zhou2021extended, ho2024optical}. INDI has proven effective in handling nonlinear systems and uncertain dynamics, making it a strong candidate for flow-divergence–based landing control.


This paper introduces a bio-inspired landing strategy for lightweight spacecraft that enables smooth and controlled touchdowns on sloped terrain using only optical flow divergence. Unlike conventional methods that rely on global estimates, our approach leverages two locally observed flow divergences to regulate thrust and attitude through an INDI-based nonlinear controller, which ensures robust flow divergence tracking despite model uncertainties. In addition, we propose a final touchdown strategy tailored for inclined surfaces: roll actuation is activated when a slope is significantly observed in the flow divergences, and ventral flow observations, defined as the ratio of horizontal velocity to height, are incorporated to mitigate lateral drift during the final phase. The method is validated in numerical simulations of landings with varying slope angles and divergence setpoints, demonstrating exponential decay of height and velocity to zero while adaptively adjusting attitude for safe and precise touchdowns. This work thus provides a lightweight, practical alternative to conventional landing strategies and represents a step toward autonomous planetary landings with resource-constrained vehicles.

The remainder of this paper is organized as follows. Section \ref{sec:dynamics_divergence} introduces the vehicle dynamics model and formulates the optical flow divergence used for landing control, including the classical constant flow divergence strategy and its extension to multiple downward-looking cameras. Section \ref{sec:Control_Strategy} presents the proposed control approach, where an INDI framework is developed to regulate thrust and attitude using local flow divergence estimates, followed by an extension to the touchdown strategy on inclined terrain. Sections \ref{sec:exp_horizontal} and \ref{sec:exp_inclined} report numerical simulation results, including landings on horizontal and inclined surfaces with different slopes, to evaluate tracking performance and touchdown behavior. Finally, Section \ref{sec:conclusion} concludes the paper with a summary of key findings and an outlook for future work.

 \section{Vehicle Dynamics and Optical Flow Divergence}
 \label{sec:dynamics_divergence}
\begin{figure}[htbp]
 	\centering
 	\input{./schematic_diagram.tex}
 	\caption{Simplified 2D spacecraft model during landing on an inclined surface.}
 	\label{fig:diagram}
 \end{figure}
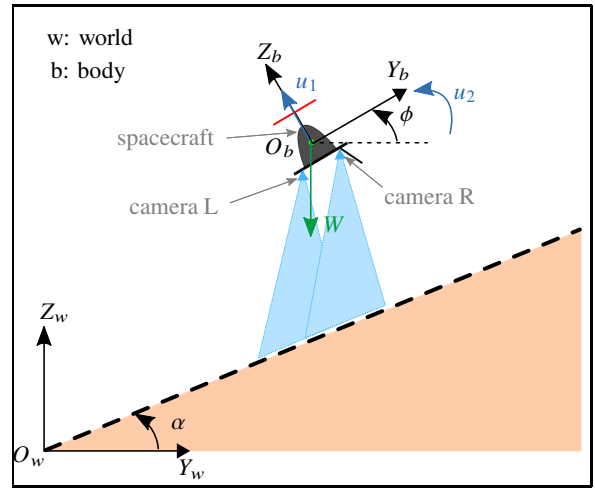

 For this preliminary study, we consider the simplified 2D dynamics of a spacecraft shown in Fig.~\ref{fig:diagram}, defined in the world reference frame ($O^wY^wZ^w$) and the body-fixed reference frame ($O^bY^bZ^b$). The translational and rotational dynamics are given by:

 \begin{equation}
 \begin{split}
     \Ddot{Y} &= -\frac{u_1}{m}\cdot sin(\phi), \\
     \Ddot{Z} &= \frac{u_1}{m}\cdot cos(\phi) - g,\\
     \Ddot{\phi} &= \frac{u_2}{I_{xx}},
 \end{split}
 \label{eq:dynamics}
 \end{equation}

 \noindent where $m$ is the vehicle mass, $I_{xx}$ is the moment of inertia about the body $x$-axis, $u_1$ is the total thrust, $u_2$ is the control moment, $\phi$ is the roll angle, and $g$ is the gravitational acceleration.
 
 To illustrate how optical flow can be utilized for landing, we begin the discussion with the classical case of a single downward-looking camera attached to the origin of the vehicle's body. During descent, the optical flow observed in the camera's field of view exhibits a divergent pattern, which we define as flow divergence:
 \begin{equation}
     \vartheta = \frac{\Dot{h}}{h},
     \label{eq:div}
 \end{equation}

 \noindent where $h$ is the vehicle's height above the ground and $\Dot{h}$ its vertical velocity. A widely used control strategy is to regulate the flow divergence to a constant negative setpoint, $\vartheta = \vartheta^* < 0$, during descent. This constant flow divergence control strategy leads to exponentially decaying height, velocity, and acceleration trajectories:

 \begin{equation}
     {h}(t) = {h}_0e^{\vartheta^*t},~\Dot{h}(t) = \vartheta^* {h}_0e^{\vartheta^*t},~\Ddot{h}(t) = {\vartheta^*}^2 {h}_0e^{\vartheta^*t},
     \label{eq:const_div_landing}
 \end{equation}
 where $h_0$ is the initial height. As $t \to \infty$, the vehicle approaches the surface with simultaneously vanishing velocity and acceleration, which ensures a smooth touchdown. 

 While effective for landing on horizontal terrain, this formulation does not account for inclined surfaces. To address this, we consider a vehicle equipped with two downward-facing cameras mounted at opposite ends of its body, as shown in Fig.~\ref{fig:diagram}. We assume that both cameras always point vertically downward in the world frame, regardless of vehicle rotation. During descent, the corresponding local flow divergences observed at the left and right cameras are then defined as:

 \begin{equation}
 \begin{split}
     \vartheta_L &= \frac{\Dot{h}_L}{h_L},\\
     \vartheta_R &= \frac{\Dot{h}_R}{h_R},
 \end{split}
 \label{eq:observation}
 \end{equation}

 \noindent where the subscripts $L$ and $R$ denote the left and right cameras, respectively, with $h_L, h_R$ the corresponding heights above the ground and $\Dot{h}_L, \Dot{h}_R$ their vertical velocities. In addition, movement in the Y-axis can also be observed via ventral flow which can be expressed using Eq.~(\ref{eq:observation2}):
 \begin{equation}
 \begin{split}
     \vartheta_y &= \frac{\Dot{Y}}{h}.
 \end{split}
 \label{eq:observation2}
 \end{equation}

 The height of the vehicle's body center above the ground, $h$, together with the camera heights $h_L$ and $h_R$, can be expressed as:

 \begin{equation}
 \begin{split}
     h &= Z-Y\cdot tan (\alpha), \\
     h_L &= h - \Delta h_\phi + \Delta h_\alpha, \\
     h_R &= h + \Delta h_\phi - \Delta h_\alpha,
 \end{split}
 \label{eq:height}
 \end{equation}

 \noindent where $\Delta h_\phi = b_c\cdot sin(\phi)$, $\Delta h_\alpha=b_c\cdot cos(\phi)\cdot tan(\alpha)$, and $b_c$ is the distance from the camera to the center of the vehicle.

 By substituting $h$ into $h_L$ and $h_R$ and simplying them using small angle approximation, we obtain:

 \begin{equation}
 \begin{split}
     h_L &= Z-Y\cdot tan (\alpha) - b_c\cdot \phi + b_c\cdot tan(\alpha), \\
     h_R &= Z-Y\cdot tan (\alpha) + b_c\cdot \phi - b_c\cdot tan(\alpha).
 \end{split}
 \label{eq:height_small}
 \end{equation}

 By taking the time derivatives of Eq.~(\ref{eq:height_small}), 

 \begin{equation}
 \begin{split}
     \Dot{h_L} &= \Dot{Z}-\Dot{Y}\cdot tan (\alpha) - b_c\cdot \Dot{\phi},  \\
     \Dot{h_R} &= \Dot{Z}-\Dot{Y}\cdot tan (\alpha) + b_c\cdot \Dot{\phi}.
 \end{split}
 \label{eq:height_derivatives}
 \end{equation}

 \section{Nonlinear Landing Control Strategy}
 \label{sec:Control_Strategy}
 As presented in the previous section, the system dynamics in Eq.(\ref{eq:dynamics}) are inherently nonlinear, involve uncertain model parameters, and provide only partial state information through the flow divergence measurements in Eqs.(\ref{eq:observation}) and (\ref{eq:observation2}). To address these challenges, we employ an Incremental Nonlinear Dynamic Inversion (INDI) approach. Section~\ref{subsec:INDI} introduces the INDI framework under the assumption of full state availability. Building on this foundation, Section~\ref{subsec:INDI_Tracking} formulates a control law based on observed optical flow to enable effective and smooth landing on inclined terrain using only optical flow. Lastly, Section~\ref{subsec:INDI_Touchdown} further extends this formulation with a touchdown strategy designed to ensure safe and stable contact on inclined terrain. 

 \subsection{Incremental Nonlinear Dynamic Inversion}
 \label{subsec:INDI}
 We begin with the general form of a nonlinear input-affine system:
 \begin{align}
 \label{eq:statemodel}
 \dot{\bm{x}} &= \textbf{\textit{f}}(\bm{x}) + \textbf{\textit{g}}(\bm{x})\bm{u},\\
 \label{eq:outputmodel}
 \bm{y} &= \bm{\eta}(\bm{x}),
 \end{align}
 where $\bm{x} \in \mathcal{R}^n$ is the system state, $\bm{u} \in \mathcal{R}^m$ is the system input, $\bm{y} \in \mathcal{R}^p$ is the system output or the observations, $\textbf{\textit{f}} : \mathcal{R}^n \rightarrow \mathcal{R}^n$ and $\bm{\eta} : \mathcal{R}^n \rightarrow \mathcal{R}^p$ are smooth vector fields, and $\textbf{\textit{g}} : \mathcal{R}^n \rightarrow \mathcal{R}^{n\times m}$, whose columns are smooth vector fields. 

 The first step in the INDI framework is input–output linearization, achieved by differentiating the outputs until the control input appears explicitly \cite{sieberling2010robust, zhou2021extended, ho2024optical}. For the $l$-th output, the first derivative is
 \begin{equation}
 \label{eq:firstOrderDifferentiation}
 \begin{split}
 \dot{y}_{l} 
 = \frac{d \eta_{l}(\bm{x})}{dt} 
 &= \frac{\partial \eta_{l}(\bm{x})}{\partial \bm{x}} \frac{d\bm{x}}{dt} \\
 &= \nabla \eta_{l}(\bm{x}) [\textbf{\textit{f}}(\bm{x}) + \textbf{\textit{g}} (\bm{x})\bm{u}]\\
 &= L_f \eta_{l} (\bm{x}) + L_g \eta_{l}(\bm{x}) \bm{u},
 \end{split}
 \end{equation}

 \noindent where $\nabla$ denotes the Jacobian operator, and $L_f \eta_{l}(\bm{x}) = \nabla \eta_{l}(\bm{x}) \textbf{\textit{f}}(\bm{x})$ is the Lie derivative along $\textbf{\textit{f}}(\bm{x})$ \cite{slotine1991applied}. If $L_g \eta_{l}(\bm{x}) \neq 0$, the system is input-affine at first order. Otherwise, further differentiation is required until:
 \begin{equation}
 \label{eq:PthOrderInputAffine}
 \begin{split}
 y_{l}^{(r_{l})} 
 = L_f^{r_{l}} \eta_{l}(\bm{x}) + L_g L_f^{r_{l}-1}\eta_{l}(\bm{x}) \bm{u},
 \end{split}
 \end{equation}
 where $r_{l}\geq 1$ is the relative degree of the $l$-th output. The total relative degree is $r = \sum_{l=1}^{p} r_{l}$, bounded by $p \leq r \leq n$.

 In compact form, the input–output dynamics can be written as:
 \begin{equation}
 \begin{split}
 \label{eq:MIMOinputAffineSymbol}
 \bm{y}^{(\bm{r})} 
 = L_f ^ {\bm{r}} \bm{\eta}(\bm{x}) 
 + L_g L_f^{\bm{r}-1}\bm{\eta}(\bm{x}) \bm{u}.
 \end{split}
 \end{equation}
 This form enables cancellation of nonlinearities and the design of desired linear dynamics \cite{slotine1991applied}. A virtual input $\bm{\nu}$ is introduced to yield a linear outer-loop relation:
 \begin{equation}
 \label{eq:PthOrderLinearLoop}
 \bm{y}^{(\bm{r})} = \bm{\nu},
 \end{equation}
 while the nonlinear inner-loop dynamics remain
 \begin{equation}
 \label{eq:PthOrderNonlinearLoop}
 \bm{\nu} = L_f^{\bm{r}} \bm{\eta}(\bm{x}) + L_g L_f^{\bm{r}-1}\bm{\eta}(\bm{x}) \bm{u}.
 \end{equation}
 The physical control input $\bm{u}$ can then be recovered by inversion of Eq.~(\ref{eq:PthOrderNonlinearLoop}):
 \begin{equation}
 \label{eq:PthOrderInversion}
 \bm{u} = L_g L_f^{\bm{r}-1}\bm{\eta}(\bm{x})^{-1} [\bm{\nu} - L_f^{\bm{r}} \bm{\eta}(\bm{x})],
 \end{equation}
 where $\bm{\nu}$ is typically selected via linear feedback in the outer loop, such as PID or polynomial pole placement, to guarantee exponentially stable dynamics. 

 However, the above NDI approach relies on the exact cancellation of both state-related terms and input-related terms, which is generally impractical, particularly in optical flow–based control scenarios. To reduce model dependency and improve stability, the INDI approach will be employed by approximating Eq. (\ref{eq:MIMOinputAffineSymbol}) using the first-order Taylor series expansion around $(\bm{x}_0,\bm{u}_0)$: 
 \begin{equation}
 \label{eq:taylorIO}
 \bm{y}^{(\bm{r})} (t) 
 \approx  \bm{y}^{(\bm{r})} _0 + F(\bm{x}_0, \bm{u}_0) \Delta \bm{x}(t) + G(\bm{x}_0, \bm{u}_0) \Delta \bm{u}(t),
 \end{equation}
 where $F(\bm{x}_0, \bm{u}_0)= $ $\frac{\partial [L_f ^ {\bm{r}} \bm{\eta}(\bm{x}) + L_g L_f^{\bm{r}-1}\bm{\eta}(\bm{x}) \bm{u}] }{\partial \bm{x}} |_{\bm{x}_0, \bm{u}_0} $ $\in \mathcal{R}^{p \times n}$, and $G(\bm{x}_0, \bm{u}_0)= $ $L_g L_f^{\bm{r}-1}\bm{\eta}(\bm{x}) |_{\bm{x}_0} $ $\in \mathcal{R}^{p \times m}$. 
 This relation can be simplified under a time-scale separation assumption, which holds when changes in the control input occur significantly faster than changes in the system state with high sampling frequencies and instantaneous control effects: 
 \begin{equation}
 \label{eq:taylorIOTSr}
 \bm{y}^{(\bm{r})} (t) 
 \approx  \bm{y}^{(\bm{r})} _0 + L_g L_f^{\bm{r}-1}\bm{\eta}(\bm{x}_0) \Delta \bm{u}(t),
 \end{equation}
 Note that, for non-affine input-output relations or unknown dynamics, INDI can also be extended via data-driven approaches \cite{zhou2021extended, ho2024optical}. 
 Based on the linearized input–output model, the control increment is then obtained by substituting the differentiated output with the virtual input $\bm{y}^{(\bm{r})} =\bm{\nu}$ and inverting Eq. (\ref{eq:taylorIOTSr}):
 \begin{equation}
 \label{eq:INDIcontrolOutputIncre}
 \Delta \bm{u}(t) =  L_g L_f^{\bm{r}-1}\bm{\eta}(\bm{x})^{-1} [\bm{\nu} - \bm{y}^{(\bm{r})} _0]. 
 \end{equation}
 Finally, when the total relative degree $r < n$, stability of the internal dynamics must be ensured to guarantee effective control.

 \subsection{Incremental Nonlinear Dynamic Inversion for Flow Divergence Tracking Control}
 \label{subsec:INDI_Tracking}
 To design an INDI controller for landing on inclined terrain, we first define the state vector as $\textbf{\textit{x}}=[Y,Z,\phi,\Dot{Y},\Dot{Z},\Dot{\phi}]^T$, the control input as $\textbf{\textit{u}}=[u_1,u_2]^T$, and the system output as $\textbf{\textit{y}}=[y_1,y_2]^T$, as given in Eq.~(\ref{eq:output}). The outputs are selected to capture the symmetric and differential components of the flow divergence measured by the two cameras:
 \begin{equation}
 \begin{split}
     y_1 &= \frac{\vartheta_R+\vartheta_L}{2},\\
     y_2 &= \vartheta_R-\vartheta_L.
 \end{split}
 \label{eq:output}
 \end{equation}
 The first output $y_1$ corresponds to the average flow divergence, which governs vertical thrust control, while $y_2$ captures the flow divergence difference, which reflects terrain inclination and guides attitude control.

 The time derivatives of the flow divergences are obtained as:
 \begin{equation}
 \begin{split}
     \Dot{\vartheta}_R &= \frac{\Ddot{h}_R}{h_R} - \frac{\Dot{h}_R^2}{h_R^2},\\
     \Dot{\vartheta}_L &= \frac{\Ddot{h}_L}{h_L} - \frac{\Dot{h}_L^2}{h_L^2}.
     \label{eq:div_derivation}
 \end{split}
 \end{equation}

 We first solve for the first and second derivatives of the right camera height:
 \begin{equation}
 \begin{split}
     \Dot{h}_R &= \Dot{Z} - \Dot{Y}\cdot tan(\alpha) + b_c\cdot \Dot{\phi},\\
     \Ddot{h}_R &= \Ddot{Z} - \Ddot{Y}\cdot tan(\alpha) + b_c\cdot \Ddot{\phi}.
     \label{eq:height_derivation_Right}
 \end{split}
 \end{equation}
 By substituting the vehicle dynamics from Eq.~(\ref{eq:dynamics}), we obtain:
 \begin{equation}
 \begin{split}
     \Ddot{h}_R &= -g+\frac{u_1}{m}\cdot cos(\phi)+\frac{u_1}{m}\cdot sin(\phi)tan(\alpha)+b_c\cdot \frac{u_2}{I_{xx}}\\
     &\approx -g + \frac{1+\phi\cdot tan(\alpha)}{m}\cdot u_1 + \frac{b_c}{I_{xx}}\cdot u_2.
     \label{eq:height_derivation_Right_dy}
 \end{split}
 \end{equation}
 Similarly, for the left camera,
 \begin{equation}
 \begin{split}
     \Dot{h}_L &= \Dot{Z} - \Dot{Y}\cdot tan(\alpha) - b_c\cdot \Dot{\phi},\\
     \Ddot{h}_L &= \Ddot{Z} - \Ddot{Y}\cdot tan(\alpha) - b_c\cdot \Ddot{\phi}\\
     &\approx -g + \frac{1+\phi\cdot tan(\alpha)}{m}\cdot u_1 - \frac{b_c}{I_{xx}}\cdot u_2.
     \label{eq:height_derivation_Left_dy}
 \end{split}
 \end{equation}
 When $b_c\ll h$, $h_L\approx h_R$, differentiating the outputs $y_1$ and $y_2$ yields approximate affine relations with respect to the control inputs:
 \begin{equation}
 \begin{split}
     \Dot{y}_1 &= \frac{\Dot{\vartheta}_R + \Dot{\vartheta}_L}{2} \\
     &= \frac{\Ddot{h}_R}{2 h_R} + \frac{\Ddot{h}_L}{2 h_L} + \zeta_1(\bm{x})\\
     & \approx \frac{1+\phi\cdot tan(\alpha)}{m\cdot [Z-Y\cdot tan(\alpha)]}\cdot u_1 - \frac{g}{Z-Y\cdot tan(\alpha)}+ \zeta_1(\bm{x}),\\
     \Dot{y}_2 &= \Dot{\vartheta}_R - \Dot{\vartheta}_L \\
     &= \frac{\Ddot{h}_R}{h_R} - \frac{\Ddot{h}_L}{h_L} + \zeta_2(\bm{x})\\
     & \approx \frac{2b_c}{I_{xx}\cdot [Z-Y\cdot tan(\alpha)]}\cdot u_2+ \zeta_2(\bm{x}),
     \label{eq:y1_derivative}
 \end{split}
 \end{equation}
 where $\zeta_1(\bm{x})=- \frac{\Dot{h}_R^2}{2 h_R^2} - \frac{\Dot{h}_L^2}{2 h_L^2}$ and $\zeta_2(\bm{x})=- \frac{\Dot{h}_R^2}{h_R^2} + \frac{\Dot{h}_L^2}{h_L^2}$ are only functions of states.

In line with the INDI formulation, we introduce two virtual control inputs in the outer-loop: one for tracking the desired flow divergence $\vartheta^*$: 
 \begin{equation}
     \nu_1 = \Dot{y}_1 = k_1\cdot (\vartheta^*-\frac{\vartheta_R+\vartheta_L}{2}) = k_1\cdot (\vartheta^*-y_1),
 \end{equation}
 and another for stabilizing the difference of local flow divergences at zero:
  \begin{equation}
 		\nu_2 = \Dot{y}_2 = k_2\cdot(\vartheta_R-\vartheta_L) = k_2\cdot (-y_2),
 		\label{eq:virtual_in2}
 \end{equation}
 \noindent with proportional gains $k_1, k_2 > 0$. 
 The inner-loop dynamics become:
 \begin{equation}
 \begin{split}
 \nu_1 &= \Dot{y}_1 + \frac{1+\phi\cdot tan(\alpha)}{m\cdot [Z-Y\cdot tan(\alpha)]}\cdot \Delta u_1,\\
 \nu_2 &= \Dot{y}_2 + \frac{2b_c}{I_{xx}\cdot [Z-Y\cdot tan(\alpha)]} \cdot \Delta u_2,\\
 \end{split}
 \end{equation}
 Inverting the dynamics to find the control increment:
 \begin{equation}
 \begin{split}
     \Delta u_1 &= \frac{m\cdot [Z-Y\cdot tan(\alpha)]}{1+\phi\cdot tan(\alpha)}\cdot (\nu_1 -\Dot{y}_1),\\
     \Delta u_2 &= \frac{I_{xx}\cdot [Z-Y\cdot tan(\alpha)]}{2b_c}\cdot  (\nu_2 -\Dot{y}_2).
     \label{eq:delta_virtual_in2}
 \end{split}
 \end{equation}
 Finally, the physical inputs are updated incrementally as:
 \begin{equation}
 \begin{split}
     u_{1,k} &= u_{1, k-1} + \Delta u_1,\\
     u_{2,k} &= u_{2, k-1} + \Delta u_2.
     \label{eq:control_input}
 \end{split}
 \end{equation}

 \subsection{Final Touchdown Strategy on Inclined Surfaces}
 \label{subsec:INDI_Touchdown}
 The INDI controller enables robust flow divergence tracking during descent. However, additional measures are required to guarantee a controlled final touchdown on inclined surfaces. In particular, the roll control input $u_2$ is not continuously active during the descent phase; instead, it is engaged only when the differential flow divergence $y_2$ exceeds a prescribed threshold $\epsilon_y$. This event indicates the imminent presence of a surface inclination, and roll actuation is then applied to align the vehicle with the slope.

 Furthermore, because the vehicle is underactuated, excessive roll adjustments during touchdown can induce undesirable lateral drift from the initial landing position. To mitigate this effect, we introduce an additional observation based on ventral flow $\vartheta_y$, denoted as $y_3$, which compensates position drift during the final phase. This observation and its time derivative can be expressed as:
 \begin{equation}
 \begin{split}
     y_3 &= \vartheta_y = \frac{\Dot{Y}}{h},\\
     \Dot{y}_3 &= \frac{\Ddot{Y}}{h} - \frac{\Dot{Y}}{h^2}\cdot \Dot{h} \\
     &= \frac{-sin(\phi)}{m\cdot [Z-Y\cdot tan(\alpha)]}\cdot u_1  + \zeta_3(\bm{x}),
     \label{eq:y3}
 \end{split}
 \end{equation}
 where $\zeta_3(\bm{x}) = - \frac{\Dot{Y} \Dot{h}}{h^2}$ is a function only of states.

 The ventral flow $y_3$ is then incorporated into the thrust channel by deriving an incremental correction for $u_1$. Analogous to the derivation of $\Delta u_1$ for $y_1$, we define the virtual input:
 \begin{equation}
     \nu_3 = \Dot{y}_3 = k_3 \cdot (-\vartheta_y) = k_3 \cdot (-y_3),
     \label{eq:virtual_in3}    
 \end{equation}
 \noindent with its proportional gain $k_3>0$. 
 It yields an additional contribution to the thrust command increment, denoted as $\Delta u_1'$:
 \begin{equation}
 \begin{split}
     \nu_3 &=  \Dot{y}_3 + \frac{-sin(\phi)}{m\cdot [Z-Y\cdot tan(\alpha)]}\cdot \Delta u_1',\\
     \Delta u_1' &= \frac{m\cdot [Z-Y\cdot tan(\alpha)]}{-sin(\phi)}\cdot (\nu_3 -  \Dot{y}_3).
     \label{eq:delta_in3}
 \end{split}
 \end{equation}
 The final thrust input is then updated as:
 \begin{equation}
     u_{1,k} = u_{1, k-1} + \Delta u_1 + \Delta u_1',
     \label{eq:control_input3}
 \end{equation}
 \noindent where the corresponding corrective increment is incorporated into the thrust command $u_1$, thereby stabilizing the vehicle's ground contact point while maintaining a soft touchdown on the slope.

 This combined strategy ensures that attitude regulation for inclined-surface alignment is only activated when necessary, while position stability is preserved through an augmented thrust control law.

 \section{Landing Experiments on Horizontal Terrain}
 \label{sec:exp_horizontal}
 This section evaluates the tracking performance of the proposed INDI controller during landings on a horizontal surface. The experiments were conducted using different flow divergence setpoints, namely -0.1, -0.2, and -0.3 rad/s. Larger negative setpoints correspond to more aggressive descent dynamics, which makes the tracking problem increasingly challenging.

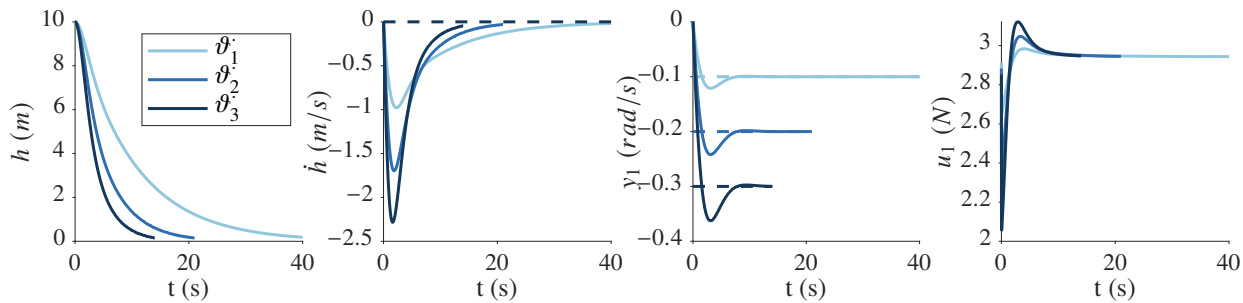
\begin{figure*}[htbp]
 	\centering
 	\input{./landing_diff_div.tex}
 	\caption{Landing experiments on horizontal terrain with flow divergence setpoints of $\vartheta_{1}^{*}$ = -0.1,  $\vartheta_{2}^{*}$ = -0.2,  $\vartheta_{3}^{*}$ = -0.3 rad/s. Plots (left to right) show the height $h$, vertical velocity $\dot{h}$, observed flow divergence $y_1$, and control input $u_1$ of the vehicle. The INDI controller achieves accurate tracking of the flow divergence setpoints, resulting in the expected exponential decay of height and velocity, with initial acceleration followed by deceleration toward touchdown.}
 	\label{fig:landing_div}
 \end{figure*}

 As shown in Fig.~\ref{fig:landing_div}, the vehicle consistently follows the expected exponential decay behavior: the height and vertical velocity both decrease smooth to zero at the same time, with the velocity profile exhibiting an initial acceleration followed by a deceleration toward touchdown. Importantly, the controller is able to accurately track the desired flow divergence across all tested setpoints. Even for the most aggressive case of -0.3~rad/s, where nonlinear effects become more pronounced, the INDI controller maintains stable tracking.

 These results demonstrate that the proposed method effectively compensates for the nonlinearities in the observation model and provides accurate flow divergence tracking. Table~\ref{tab:rmse_tracking} further quantifies this performance by comparing the Root Mean Square Errors (RMSEs) of the INDI controller with a fine-tuned classical PID controller. It can be seen that the INDI approach achieves consistently lower RMSEs across all tested flow divergence setpoints. Although RMSEs increase for larger setpoints for both controllers due to the more challenging descent dynamics, the INDI controller maintains superior accuracy, ensuring smooth and reliable autonomous landings on horizontal terrain.
 
 \begin{table}[H]
 	\caption{Landing performance on horizontal terrain using INDI and PID controllers, evaluated with the Root Mean Square Errors (RMSEs) of various $\vartheta^*$ tracking.}
	\label{tab:rmse_tracking}
 	\centering
 	\begin{tabular}{c|ccc}
 		\hline
		\multirow{2}{*}{\textbf{Controller}} & \multicolumn{3}{c}{\textbf{RMSE of $\vartheta^*$ Tracking (rad/s)}} \\
		\cline{2-4}
		& -0.1 & -0.2 & -0.3 \\
 		\hline
 		INDI & 0.0112 & 0.0310 & 0.0569 \\
 		PID  & 0.0352 & 0.0541 & 0.0735 \\ 
		\hline
 	\end{tabular}
 \end{table}


 \section{Landing Experiments on Inclined Terrain}
 \label{sec:exp_inclined}
\begin{figure*}[htbp]
 	\centering
 	\input{./landing_slope2_2.tex}
 	\caption{Landing experiments on inclined terrain with slopes of $10^\circ$, $20^\circ$, and $30^\circ$ (left to right). Plots show the vehicle's height $h$, roll angle $\phi$, flow divergence difference $y_2$, and control inputs $u_1$ and $u_2$. The results demonstrate exponential decay of height, activation of roll alignment near touchdown as $y_2$ is regulated to zero, and coordinated thrust and attitude control for stable landings across all slopes.}
 	\label{fig:landing_slope}
 \end{figure*}
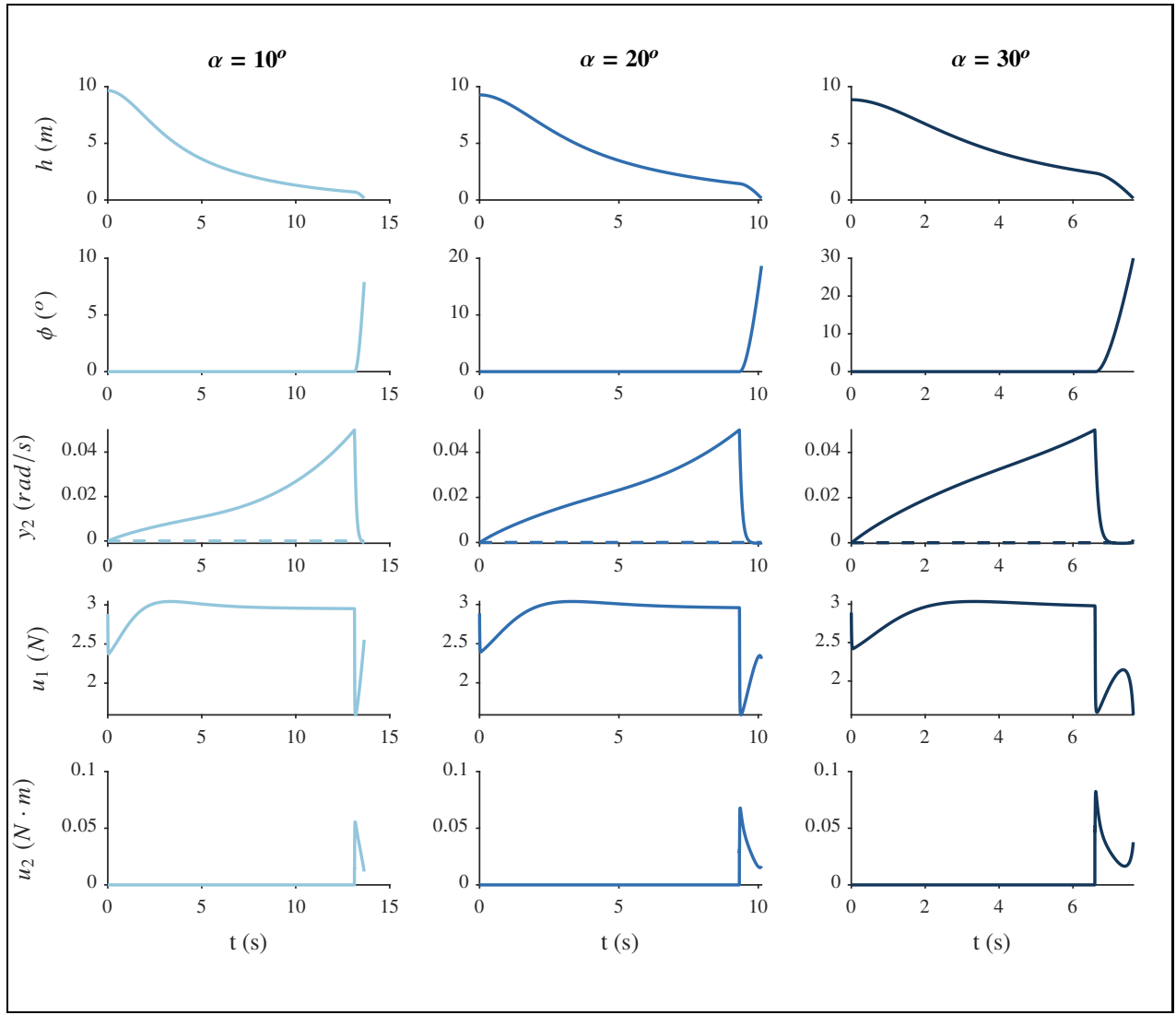

 This section investigates the performance of the proposed INDI controller during landings on inclined surfaces. To focus on the effect of terrain inclination, we fixed the flow divergence setpoint at $-0.2$ rad/s, chosen as a representative value from the previous experiments. Experiments were conducted on slopes of $10^\circ$, $20^\circ$, and $30^\circ$. The results are summarized in Fig.~\ref{fig:landing_slope}, which shows the evolution of the vehicle's height $h$, roll angle $\phi$, flow divergence observation $y_2$, and control inputs $u_1$ and $u_2$.

 
 For all tested slopes, the vertical motion exhibits the expected exponential decay prior to touchdown, as clearly shown in the height plots in Fig.~\ref{fig:landing_slope}, similar to the landing case on horizontal terrain. This descent is governed primarily by the thrust input $u_1$, which regulates the average flow divergence and ensures smooth vertical dynamics. Near touchdown, the roll angle begins to adjust toward the slope inclination, as evident from the roll angle plots in the same figure. This alignment is triggered when $y_2$ becomes significant, indicating a large difference between the left and right flow divergences, and is then driven toward zero to complete the attitude adjustment. The corresponding control input $u_2$ shows the activation of roll control during this phase. Simultaneously, $u_1$ adapts to mitigate the lateral drift induced by rolling. The alignment occurs during a slow descent phase, which provides a safe and stable touchdown on the inclined surface. 
 
\begin{figure*}[htbp]
 	\centering
 	\input{./anime.tex}
 	\input{./anime_x_comp.tex}
 	\caption{Landing trajectories on inclined terrain with slopes of (a) $10^\circ$, (b) $20^\circ$, and (c) $30^\circ$ using the proposed method, and a final touchdown without drift compensation (rightmost). (a-c) Vehicle motion is shown in the $(Z,Y)$ plane with snapshots overlaid at fixed time intervals. The sequence highlights the initial acceleration (large spacing between snapshots), gradual deceleration (smaller spacing), and final touchdown with the roll angle aligned to the slope. The rightmost case without drift compensation shows the vehicle drifting far away from the intended landing site.}
 	\label{fig:anime}
 \end{figure*}
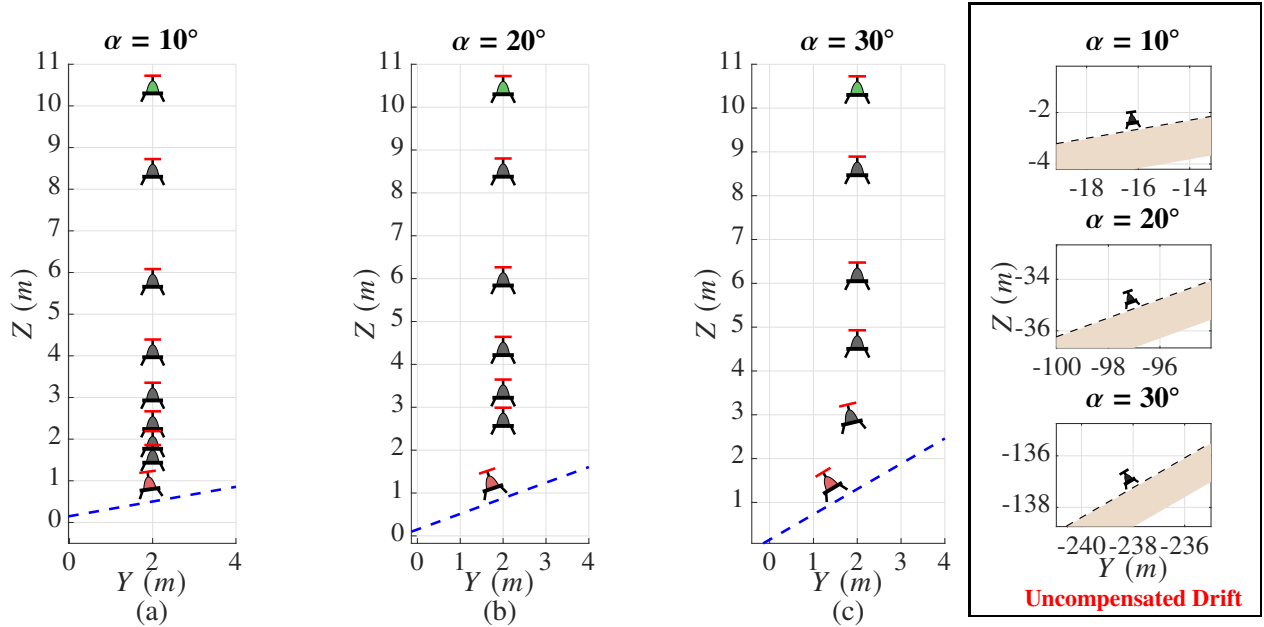

 The landing trajectories are further illustrated in Fig.~\ref{fig:anime}, which shows the evolution of the vehicle position in the ($Z$, $Y$) plane. Snapshots of the vehicle configuration are overlaid at fixed time intervals. The plots reveal the expected phases of the maneuver: an initial acceleration phase with large inter-snapshot spacing, a subsequent deceleration phase with progressively smaller spacing, and finally a stable touchdown with the vehicle roll angle aligned with the slope. In these cases, lateral drift is effectively compensated by our method. In contrast, the rightmost plot in Fig.~\ref{fig:anime} shows landings without drift compensation for the same slopes. Here, the vehicle drifts increasingly farther from the intended landing site, highlighting the importance of the proposed compensation strategy.

 Furthermore, we also compare it with a fine-tuned PID controller for landings on inclined terrain. Table~\ref{tab:inclined_comparison} summarizes the results in terms of RMSE of $y_2$ regulation, final roll angle during touchdown $\phi_f$, and lateral drift $Y_\text{drift}$. The RMSE of INDI is consistently smaller than that of the PID controller across all tested slopes. Although the PID controller achieves roll angles closer to the slope for the first two inclinations, it exhibits overshoot for the $30^\circ$ case with a larger deviation. Most importantly, INDI provides superior drift compensation, maintaining the vehicle closer to the intended landing site.
 
  \begin{table*}[htbp]
	\centering
	\caption{Landing performance on inclined terrain for various $\alpha$ using INDI and PID controllers (with drift compensation), evaluated with the RMSE of $y_2$ regulation (rad/s), final roll angle $\phi_f$ ($^\circ$), and lateral drift $Y_\text{drift}$ (m).}
	\label{tab:inclined_comparison}
	\begin{tabular}{c|ccc|ccc|ccc}
		\hline
		\multirow{3}{*}{Controller} & \multicolumn{3}{c|}{$\alpha=10^\circ$} & \multicolumn{3}{c|}{$\alpha=20^\circ$} & \multicolumn{3}{c}{$\alpha=30^\circ$} \\
		\cline{2-10}
		& RMSE & $\phi_f$ & $Y_\text{drift}$ & RMSE & $\phi_f$ & $Y_\text{drift}$ & RMSE & $\phi_f$ & $Y_\text{drift}$\\
		\hline
		INDI & 0.0150 & 7.9 & -0.02 & 0.0119 & 18.6 & -0.14 & 0.0104 & 31.0 & -0.41 \\
		PID  & 0.1413 & 9.4 & -0.43 & 0.1216 & 20.1 & -1.05 & 0.1010 & 32.2 & -1.75 \\
		\hline
 	\end{tabular}
 \end{table*}
 
  Overall, these results validate that the INDI-based controller enables robust landings on inclined terrain by effectively coordinating vertical descent control through $u_1$ and attitude alignment through $u_2$. The combination of flow-divergence-based regulation and adaptive attitude control provides smooth touchdown performance with reduced lateral drift even under steep slope conditions.

\section{Conclusions}
\label{sec:conclusion}
This paper presented a bio-inspired landing strategy for lightweight spacecraft that leverages optical flow divergence observables from two cameras and an Incremental Nonlinear Dynamic Inversion (INDI) controller to achieve smooth, controlled touchdowns on inclined terrain. By regulating the average divergence for thrust control and their difference for attitude alignment, the proposed method enables exponential decay of height and velocity on horizontal surfaces and robust slope alignment during inclined landings. Simulation results demonstrated successful landings with stable performance across different flow divergence setpoints and terrain slopes, which highlights the controller's ability to handle nonlinearities with only optical flow observables. This approach offers a practical alternative to conventional landing methods and provides a foundation for extending bio-inspired strategies to future autonomous planetary landing missions. Future work will focus on generalizing the framework to 3D dynamics, incorporating sensor noise and delays, exploring integration with vision-based terrain perception, and validating the method in real-world flight experiments.

\section*{Acknowledgements}
The authors would like to thank Malaysian Ministry of Higher Education (MOHE) for providing the Fundamental Research Grant Scheme (FRGS) (Grant number: FRGS/1/2024/TK04/USM/02/3) for conducting this research.


\section*{References}
\printbibliography[heading=none]

\end{document}

%% file: schematic_diagram.tex
%
%
\providecommand\matlabfragNegTickNoWidth{\makebox[0pt][r]{\ensuremath{-}}}%
\newcommand{\tsize}{0.9}
%
\psfrag{O}[c][c][\tsize]{\color[rgb]{0,0,0}\setlength{\tabcolsep}{0pt}\begin{tabular}{c}$O_w$\end{tabular}}%
\psfrag{Y}[c][c][\tsize]{\color[rgb]{0,0,0}\setlength{\tabcolsep}{0pt}\begin{tabular}{c}$Y_w$\end{tabular}}%
\psfrag{Z}[c][c][\tsize]{\color[rgb]{0,0,0}\setlength{\tabcolsep}{0pt}\begin{tabular}{c}$Z_w$\end{tabular}}%
\psfrag{d}[c][c][\tsize]{\color[rgb]{0,0,0}\setlength{\tabcolsep}{0pt}\begin{tabular}{c}$O_b$\end{tabular}}%
\psfrag{m}[c][c][\tsize]{\color[rgb]{0,0,0}\setlength{\tabcolsep}{0pt}\begin{tabular}{c}$Z_b$\end{tabular}}%
\psfrag{n}[c][c][\tsize]{\color[rgb]{0,0,0}\setlength{\tabcolsep}{0pt}\begin{tabular}{c}$Y_b$\end{tabular}}%
\psfrag{i}[c][c][\tsize]{\color[rgb]{0,0,0}\setlength{\tabcolsep}{0pt}\begin{tabular}{c}w: world\end{tabular}}%
\psfrag{j}[c][c][\tsize]{\color[rgb]{0,0,0}\setlength{\tabcolsep}{0pt}\begin{tabular}{c}b: body\end{tabular}}%
\psfrag{w}[c][c][\tsize]{\color[rgb]{0,0.61,0.25}\setlength{\tabcolsep}{0pt}\begin{tabular}{c}$W$\end{tabular}}%
\psfrag{u}[c][c][\tsize]{\color[rgb]{0.19,0.43,0.68}\setlength{\tabcolsep}{0pt}\begin{tabular}{c}$u_1$\end{tabular}}%
\psfrag{t}[c][c][\tsize]{\color[rgb]{0.19,0.43,0.68}\setlength{\tabcolsep}{0pt}\begin{tabular}{c}$u_2$\end{tabular}}%
\psfrag{p}[c][c][\tsize]{\color[rgb]{0,0,0}\setlength{\tabcolsep}{0pt}\begin{tabular}{c}$\phi$\end{tabular}}%
\psfrag{a}[c][c][\tsize]{\color[rgb]{0,0,0}\setlength{\tabcolsep}{0pt}\begin{tabular}{c}$\alpha$\end{tabular}}%
\psfrag{c}[c][c][\tsize]{\color[rgb]{0.5,0.5,0.5}\setlength{\tabcolsep}{0pt}\begin{tabular}{c}camera L\end{tabular}}%
\psfrag{k}[c][c][\tsize]{\color[rgb]{0.5,0.5,0.5}\setlength{\tabcolsep}{0pt}\begin{tabular}{c}camera R\end{tabular}}%
\psfrag{s}[c][c][\tsize]{\color[rgb]{0.5,0.5,0.5}\setlength{\tabcolsep}{0pt}\begin{tabular}{c}spacecraft\end{tabular}}%
%
%
\fbox{\includegraphics[trim=0 0 0 0, clip, width =0.45\textwidth ]{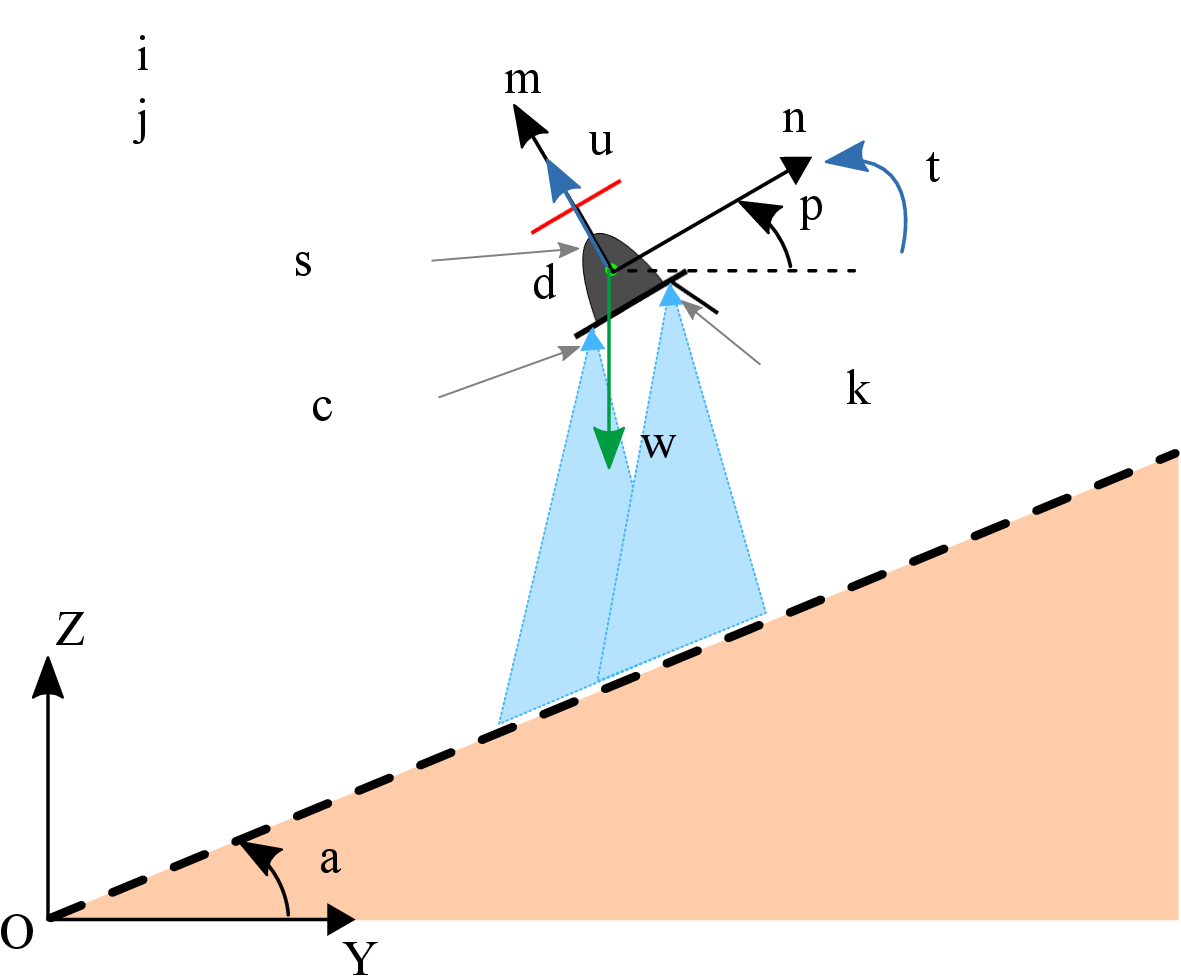}}

%% file: landing_diff_div.tex
%
\providecommand\matlabfragNegTickNoWidth{\makebox[0pt][r]{\ensuremath{-}}}%
%
%
\providecommand\matlabtextA{\color[rgb]{0.000,0.000,0.000}\fontsize{10.00}{10.00}\selectfont\strut}%
\psfrag{00000}[cl][cl]{\matlabtextA $\vartheta_{1}^{*}$}%
\psfrag{00001}[cl][cl]{\matlabtextA $\vartheta_{2}^{*}$}%
\psfrag{00002}[cl][cl]{\matlabtextA $\vartheta_{3}^{*}$}%
\providecommand\matlabtextB{\color[rgb]{0.150,0.150,0.150}\fontsize{9.90}{9.90}\selectfont\strut}%
\psfrag{038}[tc][tc]{\matlabtextB t~(s)}%
\psfrag{039}[bc][bc]{\matlabtextB $u_1~(N)$}%
\psfrag{040}[tc][tc]{\matlabtextB t~(s)}%
\psfrag{041}[bc][bc]{\matlabtextB $y_1~(rad/s)$}%
\psfrag{042}[tc][tc]{\matlabtextB t~(s)}%
\psfrag{043}[bc][bc]{\matlabtextB $\dot{h}~(m/s)$}%
\psfrag{044}[tc][tc]{\matlabtextB t~(s)}%
\psfrag{045}[bc][bc]{\matlabtextB $h~(m)$}%
%
%
%
\providecommand\matlabtextC{\color[rgb]{0.150,0.150,0.150}\fontsize{9.00}{9.00}\selectfont\strut}%
\psfrag{003}[ct][ct]{\matlabtextC $0$}%
\psfrag{004}[ct][ct]{\matlabtextC $20$}%
\psfrag{005}[ct][ct]{\matlabtextC $40$}%
\psfrag{012}[ct][ct]{\matlabtextC $0$}%
\psfrag{013}[ct][ct]{\matlabtextC $20$}%
\psfrag{014}[ct][ct]{\matlabtextC $40$}%
\psfrag{021}[ct][ct]{\matlabtextC $0$}%
\psfrag{022}[ct][ct]{\matlabtextC $20$}%
\psfrag{023}[ct][ct]{\matlabtextC $40$}%
\psfrag{030}[ct][ct]{\matlabtextC $0$}%
\psfrag{031}[ct][ct]{\matlabtextC $20$}%
\psfrag{032}[ct][ct]{\matlabtextC $40$}%
%
%
%
\psfrag{006}[rc][rc]{\matlabtextC $0$}%
\psfrag{007}[rc][rc]{\matlabtextC $2$}%
\psfrag{008}[rc][rc]{\matlabtextC $4$}%
\psfrag{009}[rc][rc]{\matlabtextC $6$}%
\psfrag{010}[rc][rc]{\matlabtextC $8$}%
\psfrag{011}[rc][rc]{\matlabtextC $10$}%
\psfrag{015}[rc][rc]{\matlabtextC $2$}%
\psfrag{016}[rc][rc]{\matlabtextC $2.2$}%
\psfrag{017}[rc][rc]{\matlabtextC $2.4$}%
\psfrag{018}[rc][rc]{\matlabtextC $2.6$}%
\psfrag{019}[rc][rc]{\matlabtextC $2.8$}%
\psfrag{020}[rc][rc]{\matlabtextC $3$}%
\psfrag{024}[rc][rc]{\matlabtextC $-2.5$}%
\psfrag{025}[rc][rc]{\matlabtextC $-2$}%
\psfrag{026}[rc][rc]{\matlabtextC $-1.5$}%
\psfrag{027}[rc][rc]{\matlabtextC $-1$}%
\psfrag{028}[rc][rc]{\matlabtextC $-0.5$}%
\psfrag{029}[rc][rc]{\matlabtextC $0$}%
\psfrag{033}[rc][rc]{\matlabtextC $-0.4$}%
\psfrag{034}[rc][rc]{\matlabtextC $-0.3$}%
\psfrag{035}[rc][rc]{\matlabtextC $-0.2$}%
\psfrag{036}[rc][rc]{\matlabtextC $-0.1$}%
\psfrag{037}[rc][rc]{\matlabtextC $0$}%
%
\includegraphics[trim=80 0 50 0, clip, width =1.0\textwidth ]{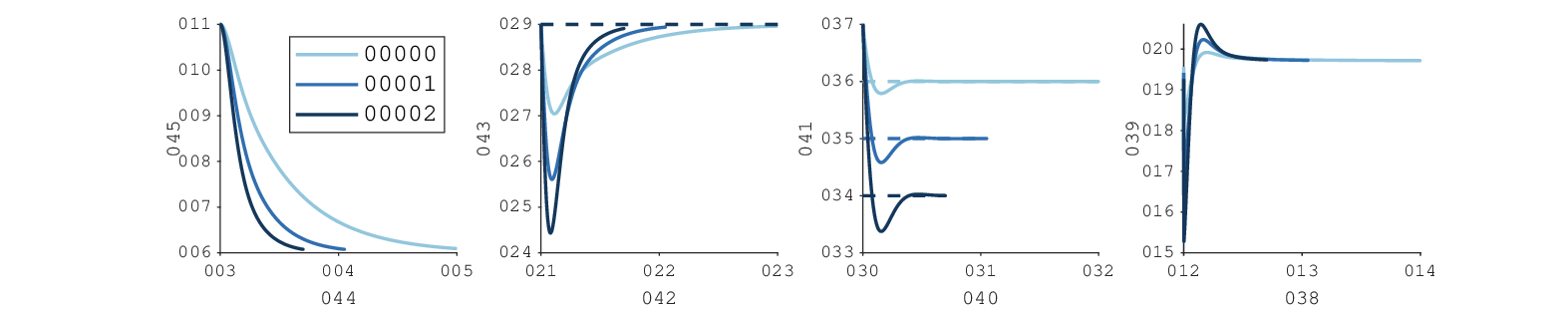}

%% file: landing_slope2_2.tex
%
\providecommand\matlabfragNegTickNoWidth{\makebox[0pt][r]{\ensuremath{-}}}%
%
%
\providecommand\matlabtextA{\color[rgb]{0.150,0.150,0.150}\fontsize{9.90}{9.90}\selectfont\strut}%
\psfrag{101}[tc][tc]{\matlabtextA t~(s)}%
\psfrag{102}[tc][tc]{\matlabtextA t~(s)}%
\psfrag{103}[tc][tc]{\matlabtextA t~(s)}%
\psfrag{104}[bc][bc]{\matlabtextA $u_2~(N\cdot m)$}%
\psfrag{105}[bc][bc]{\matlabtextA $u_1~(N)$}%
\psfrag{106}[bc][bc]{\matlabtextA $y_2~(rad/s)$}%
\psfrag{107}[bc][bc]{\matlabtextA $\phi~(^o)$}%
\psfrag{111}[bc][bc]{\matlabtextA $h~(m)$}%
\providecommand\matlabtextB{\color[rgb]{0.000,0.000,0.000}\fontsize{9.90}{9.90}\bfseries\boldmath\selectfont\strut}%
\psfrag{108}[bc][bc]{\matlabtextB $\alpha=30^o$}%
\psfrag{109}[bc][bc]{\matlabtextB $\alpha=20^o$}%
\psfrag{110}[bc][bc]{\matlabtextB $\alpha=10^o$}%
%
%
%
\providecommand\matlabtextC{\color[rgb]{0.150,0.150,0.150}\fontsize{8.00}{8.00}\selectfont\strut}%
\psfrag{000}[ct][ct]{\matlabtextC $0$}%
\psfrag{001}[ct][ct]{\matlabtextC $5$}%
\psfrag{002}[ct][ct]{\matlabtextC $10$}%
\psfrag{003}[ct][ct]{\matlabtextC $15$}%
\psfrag{007}[ct][ct]{\matlabtextC $0$}%
\psfrag{008}[ct][ct]{\matlabtextC $2$}%
\psfrag{009}[ct][ct]{\matlabtextC $4$}%
\psfrag{010}[ct][ct]{\matlabtextC $6$}%
\psfrag{014}[ct][ct]{\matlabtextC $0$}%
\psfrag{015}[ct][ct]{\matlabtextC $5$}%
\psfrag{016}[ct][ct]{\matlabtextC $10$}%
\psfrag{017}[ct][ct]{\matlabtextC $15$}%
\psfrag{021}[ct][ct]{\matlabtextC $0$}%
\psfrag{022}[ct][ct]{\matlabtextC $2$}%
\psfrag{023}[ct][ct]{\matlabtextC $4$}%
\psfrag{024}[ct][ct]{\matlabtextC $6$}%
\psfrag{028}[ct][ct]{\matlabtextC $0$}%
\psfrag{029}[ct][ct]{\matlabtextC $5$}%
\psfrag{030}[ct][ct]{\matlabtextC $10$}%
\psfrag{034}[ct][ct]{\matlabtextC $0$}%
\psfrag{035}[ct][ct]{\matlabtextC $2$}%
\psfrag{036}[ct][ct]{\matlabtextC $4$}%
\psfrag{037}[ct][ct]{\matlabtextC $6$}%
\psfrag{042}[ct][ct]{\matlabtextC $0$}%
\psfrag{043}[ct][ct]{\matlabtextC $5$}%
\psfrag{044}[ct][ct]{\matlabtextC $10$}%
\psfrag{048}[ct][ct]{\matlabtextC $0$}%
\psfrag{049}[ct][ct]{\matlabtextC $5$}%
\psfrag{050}[ct][ct]{\matlabtextC $10$}%
\psfrag{054}[ct][ct]{\matlabtextC $0$}%
\psfrag{055}[ct][ct]{\matlabtextC $5$}%
\psfrag{056}[ct][ct]{\matlabtextC $10$}%
\psfrag{057}[ct][ct]{\matlabtextC $15$}%
\psfrag{061}[ct][ct]{\matlabtextC $0$}%
\psfrag{062}[ct][ct]{\matlabtextC $2$}%
\psfrag{063}[ct][ct]{\matlabtextC $4$}%
\psfrag{064}[ct][ct]{\matlabtextC $6$}%
\psfrag{068}[ct][ct]{\matlabtextC $0$}%
\psfrag{069}[ct][ct]{\matlabtextC $5$}%
\psfrag{070}[ct][ct]{\matlabtextC $10$}%
\psfrag{074}[ct][ct]{\matlabtextC $0$}%
\psfrag{075}[ct][ct]{\matlabtextC $5$}%
\psfrag{076}[ct][ct]{\matlabtextC $10$}%
\psfrag{077}[ct][ct]{\matlabtextC $15$}%
\psfrag{081}[ct][ct]{\matlabtextC $0$}%
\psfrag{082}[ct][ct]{\matlabtextC $5$}%
\psfrag{083}[ct][ct]{\matlabtextC $10$}%
\psfrag{087}[ct][ct]{\matlabtextC $0$}%
\psfrag{088}[ct][ct]{\matlabtextC $2$}%
\psfrag{089}[ct][ct]{\matlabtextC $4$}%
\psfrag{090}[ct][ct]{\matlabtextC $6$}%
\psfrag{094}[ct][ct]{\matlabtextC $0$}%
\psfrag{095}[ct][ct]{\matlabtextC $5$}%
\psfrag{096}[ct][ct]{\matlabtextC $10$}%
\psfrag{097}[ct][ct]{\matlabtextC $15$}%
%
%
%
\psfrag{004}[rc][rc]{\matlabtextC $0$}%
\psfrag{005}[rc][rc]{\matlabtextC $0.02$}%
\psfrag{006}[rc][rc]{\matlabtextC $0.04$}%
\psfrag{011}[rc][rc]{\matlabtextC $0$}%
\psfrag{012}[rc][rc]{\matlabtextC $0.02$}%
\psfrag{013}[rc][rc]{\matlabtextC $0.04$}%
\psfrag{018}[rc][rc]{\matlabtextC $2$}%
\psfrag{019}[rc][rc]{\matlabtextC $2.5$}%
\psfrag{020}[rc][rc]{\matlabtextC $3$}%
\psfrag{025}[rc][rc]{\matlabtextC $2$}%
\psfrag{026}[rc][rc]{\matlabtextC $2.5$}%
\psfrag{027}[rc][rc]{\matlabtextC $3$}%
\psfrag{031}[rc][rc]{\matlabtextC $2$}%
\psfrag{032}[rc][rc]{\matlabtextC $2.5$}%
\psfrag{033}[rc][rc]{\matlabtextC $3$}%
\psfrag{038}[rc][rc]{\matlabtextC $0$}%
\psfrag{039}[rc][rc]{\matlabtextC $10$}%
\psfrag{040}[rc][rc]{\matlabtextC $20$}%
\psfrag{041}[rc][rc]{\matlabtextC $30$}%
\psfrag{045}[rc][rc]{\matlabtextC $0$}%
\psfrag{046}[rc][rc]{\matlabtextC $10$}%
\psfrag{047}[rc][rc]{\matlabtextC $20$}%
\psfrag{051}[rc][rc]{\matlabtextC $0$}%
\psfrag{052}[rc][rc]{\matlabtextC $0.02$}%
\psfrag{053}[rc][rc]{\matlabtextC $0.04$}%
\psfrag{058}[rc][rc]{\matlabtextC $0$}%
\psfrag{059}[rc][rc]{\matlabtextC $0.05$}%
\psfrag{060}[rc][rc]{\matlabtextC $0.1$}%
\psfrag{065}[rc][rc]{\matlabtextC $0$}%
\psfrag{066}[rc][rc]{\matlabtextC $0.05$}%
\psfrag{067}[rc][rc]{\matlabtextC $0.1$}%
\psfrag{071}[rc][rc]{\matlabtextC $0$}%
\psfrag{072}[rc][rc]{\matlabtextC $0.05$}%
\psfrag{073}[rc][rc]{\matlabtextC $0.1$}%
\psfrag{078}[rc][rc]{\matlabtextC $0$}%
\psfrag{079}[rc][rc]{\matlabtextC $5$}%
\psfrag{080}[rc][rc]{\matlabtextC $10$}%
\psfrag{084}[rc][rc]{\matlabtextC $0$}%
\psfrag{085}[rc][rc]{\matlabtextC $5$}%
\psfrag{086}[rc][rc]{\matlabtextC $10$}%
\psfrag{091}[rc][rc]{\matlabtextC $0$}%
\psfrag{092}[rc][rc]{\matlabtextC $5$}%
\psfrag{093}[rc][rc]{\matlabtextC $10$}%
\psfrag{098}[rc][rc]{\matlabtextC $0$}%
\psfrag{099}[rc][rc]{\matlabtextC $5$}%
\psfrag{100}[rc][rc]{\matlabtextC $10$}%
%
\fbox{\includegraphics[trim=30 0 30 0, clip, width =1.0\textwidth ]{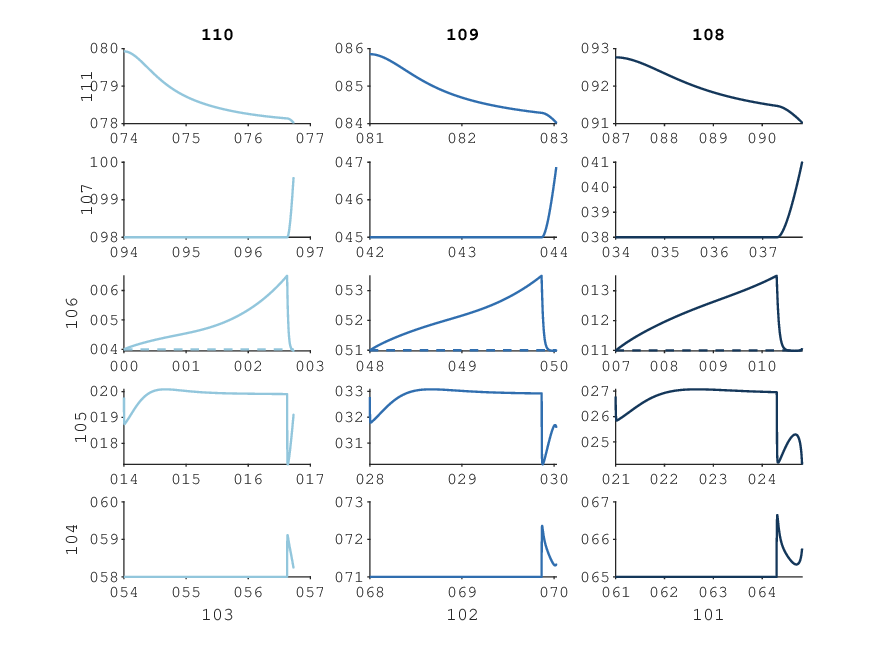}}

%% file: anime.tex
%
\providecommand\matlabfragNegTickNoWidth{\makebox[0pt][r]{\ensuremath{-}}}%
%
%
\providecommand\matlabtextA{\color[rgb]{0.150,0.150,0.150}\fontsize{11.00}{11.00}\selectfont\strut}%
\psfrag{049}[tc][tc]{\matlabtextA \begin{tabular}{c}$Y~(m)$\\ \text{(c)} \end{tabular}}%
\psfrag{050}[bc][bc]{\matlabtextA $Z~(m)$}%
\psfrag{052}[tc][tc]{\matlabtextA\begin{tabular}{c}$Y~(m)$\\ \text{(b)} \end{tabular}}%
\psfrag{053}[bc][bc]{\matlabtextA $Z~(m)$}%
\psfrag{055}[tc][tc]{\matlabtextA \begin{tabular}{c}$Y~(m)$\\ \text{(a)} \end{tabular}}%
\psfrag{056}[bc][bc]{\matlabtextA $Z~(m)$}%
\providecommand\matlabtextB{\color[rgb]{0.000,0.000,0.000}\fontsize{11.00}{11.00}\bfseries\boldmath\selectfont\strut}%
\psfrag{048}[bc][bc]{\matlabtextB $\alpha=30^\circ$}%
\psfrag{051}[bc][bc]{\matlabtextB $\alpha=20^\circ$}%
\psfrag{054}[bc][bc]{\matlabtextB $\alpha=10^\circ$}%
%
%
%
\providecommand\matlabtextC{\color[rgb]{0.150,0.150,0.150}\fontsize{10.00}{10.00}\selectfont\strut}%
\psfrag{000}[ct][ct]{\matlabtextC $0$}%
\psfrag{001}[ct][ct]{\matlabtextC $2$}%
\psfrag{002}[ct][ct]{\matlabtextC $4$}%
\psfrag{015}[ct][ct]{\matlabtextC $0$}%
\psfrag{016}[ct][ct]{\matlabtextC $1$}%
\psfrag{017}[ct][ct]{\matlabtextC $2$}%
\psfrag{018}[ct][ct]{\matlabtextC $3$}%
\psfrag{019}[ct][ct]{\matlabtextC $4$}%
\psfrag{031}[ct][ct]{\matlabtextC $0$}%
\psfrag{032}[ct][ct]{\matlabtextC $1$}%
\psfrag{033}[ct][ct]{\matlabtextC $2$}%
\psfrag{034}[ct][ct]{\matlabtextC $3$}%
\psfrag{035}[ct][ct]{\matlabtextC $4$}%
%
%
%
\psfrag{003}[rc][rc]{\matlabtextC $0$}%
\psfrag{004}[rc][rc]{\matlabtextC $1$}%
\psfrag{005}[rc][rc]{\matlabtextC $2$}%
\psfrag{006}[rc][rc]{\matlabtextC $3$}%
\psfrag{007}[rc][rc]{\matlabtextC $4$}%
\psfrag{008}[rc][rc]{\matlabtextC $5$}%
\psfrag{009}[rc][rc]{\matlabtextC $6$}%
\psfrag{010}[rc][rc]{\matlabtextC $7$}%
\psfrag{011}[rc][rc]{\matlabtextC $8$}%
\psfrag{012}[rc][rc]{\matlabtextC $9$}%
\psfrag{013}[rc][rc]{\matlabtextC $10$}%
\psfrag{014}[rc][rc]{\matlabtextC $11$}%
\psfrag{020}[rc][rc]{\matlabtextC $1$}%
\psfrag{021}[rc][rc]{\matlabtextC $2$}%
\psfrag{022}[rc][rc]{\matlabtextC $3$}%
\psfrag{023}[rc][rc]{\matlabtextC $4$}%
\psfrag{024}[rc][rc]{\matlabtextC $5$}%
\psfrag{025}[rc][rc]{\matlabtextC $6$}%
\psfrag{026}[rc][rc]{\matlabtextC $7$}%
\psfrag{027}[rc][rc]{\matlabtextC $8$}%
\psfrag{028}[rc][rc]{\matlabtextC $9$}%
\psfrag{029}[rc][rc]{\matlabtextC $10$}%
\psfrag{030}[rc][rc]{\matlabtextC $11$}%
\psfrag{036}[rc][rc]{\matlabtextC $0$}%
\psfrag{037}[rc][rc]{\matlabtextC $1$}%
\psfrag{038}[rc][rc]{\matlabtextC $2$}%
\psfrag{039}[rc][rc]{\matlabtextC $3$}%
\psfrag{040}[rc][rc]{\matlabtextC $4$}%
\psfrag{041}[rc][rc]{\matlabtextC $5$}%
\psfrag{042}[rc][rc]{\matlabtextC $6$}%
\psfrag{043}[rc][rc]{\matlabtextC $7$}%
\psfrag{044}[rc][rc]{\matlabtextC $8$}%
\psfrag{045}[rc][rc]{\matlabtextC $9$}%
\psfrag{046}[rc][rc]{\matlabtextC $10$}%
\psfrag{047}[rc][rc]{\matlabtextC $11$}%
%
\includegraphics[trim=130 0 110 0, clip, width =0.75\textwidth ]{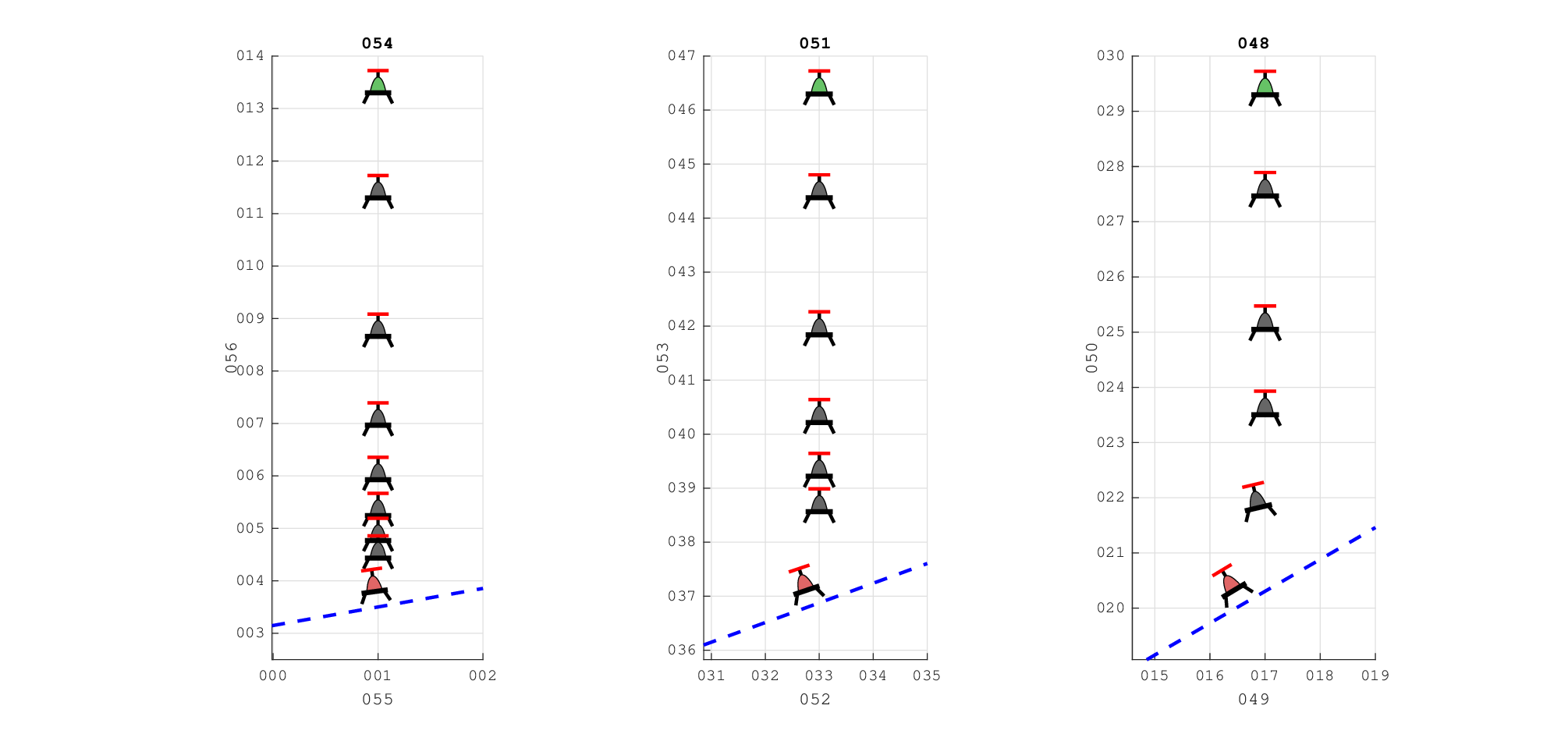} \hfill

%% file: anime_x_comp.tex
%
\providecommand\matlabfragNegTickNoWidth{\makebox[0pt][r]{\ensuremath{-}}}%
%
%
\providecommand\matlabtextA{\color[rgb]{0.150,0.150,0.150}\fontsize{8.00}{8.00}\selectfont\strut}%
\psfrag{016}[tc][tc]{\matlabtextA \begin{tabular}{c}$Y~(m)$\\ \textcolor{red}{\textbf{\small Uncompensated Drift}}\end{tabular}}%
\psfrag{017}[bc][bc]{\matlabtextA }%
\psfrag{019}[tc][tc]{\matlabtextA }%
\psfrag{020}[bc][bc]{\matlabtextA $Z~(m)$}%
\psfrag{022}[tc][tc]{\matlabtextA }%
\psfrag{023}[bc][bc]{\matlabtextA }%
\providecommand\matlabtextB{\color[rgb]{0.000,0.000,0.000}\fontsize{9.00}{9.00}\bfseries\boldmath\selectfont\strut}%
\psfrag{015}[bc][bc]{\matlabtextB $\alpha=30^\circ$}%
\psfrag{018}[bc][bc]{\matlabtextB $\alpha=20^\circ$}%
\psfrag{021}[bc][bc]{\matlabtextB $\alpha=10^\circ$}%
%
%
%
\providecommand\matlabtextC{\color[rgb]{0.150,0.150,0.150}\fontsize{6.00}{6.00}\selectfont\strut}%
\psfrag{000}[ct][ct]{\matlabtextC -100}%
\psfrag{001}[ct][ct]{\matlabtextC -98}%
\psfrag{002}[ct][ct]{\matlabtextC -96}%
\psfrag{005}[ct][ct]{\matlabtextC -18}%
\psfrag{006}[ct][ct]{\matlabtextC -16}%
\psfrag{007}[ct][ct]{\matlabtextC -14}%
\psfrag{010}[ct][ct]{\matlabtextC -240}%
\psfrag{011}[ct][ct]{\matlabtextC -238}%
\psfrag{012}[ct][ct]{\matlabtextC -236}%
%
%
%
\psfrag{003}[rc][rc]{\matlabtextC -36}%
\psfrag{004}[rc][rc]{\matlabtextC -34}%
\psfrag{008}[rc][rc]{\matlabtextC -4}%
\psfrag{009}[rc][rc]{\matlabtextC -2}%
\psfrag{013}[rc][rc]{\matlabtextC -138}%
\psfrag{014}[rc][rc]{\matlabtextC -136}%
%
\fbox{\includegraphics[trim=140 0 140 0, clip, width =0.22\textwidth ]{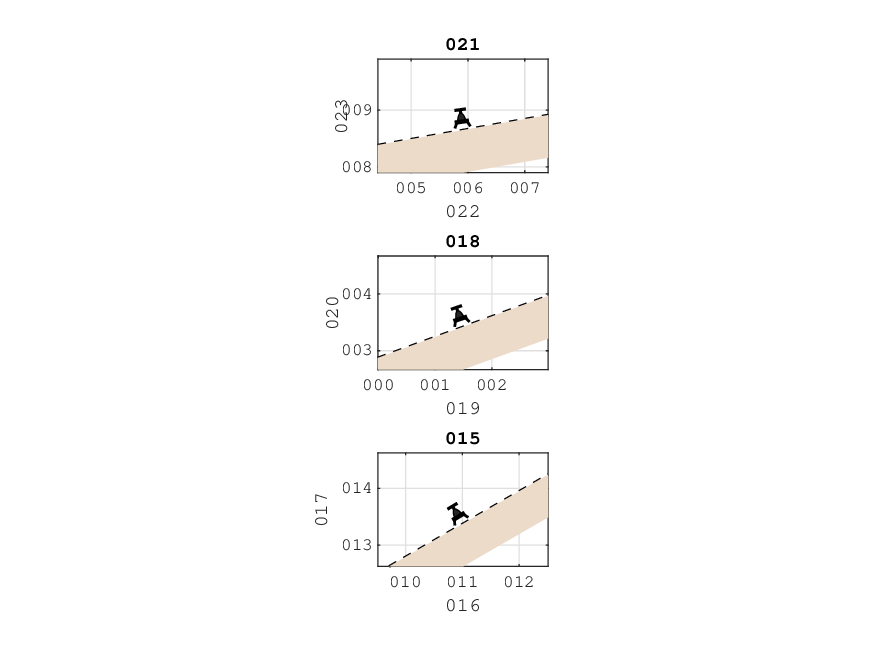}}